\documentclass[journal]{IEEEtran}

\ifCLASSINFOpdf
\else
   \usepackage[dvips]{graphicx}
\fi
\usepackage{url}

\usepackage{xcolor}
\usepackage{graphicx}
\usepackage{pifont}
\usepackage{amsmath}
\usepackage{amsfonts}
\usepackage{booktabs} % 提供 \toprule, 
\usepackage{multirow} % 支持跨行
\usepackage{hhline}   % 修复 Table I 报错：你使用了 \hhline 但没加这个包
\usepackage{makecell} % 方便在单元格内换行
\usepackage{cite}      % 用于基本的文献引用自动排序/压缩
\usepackage{hyperref} % 生成可点击的超链接，但在打印版隐藏红框

\usepackage[capitalize]{cleveref}
\crefname{figure}{Fig.}{Figs.}
\crefname{section}{Section}{Sections}
\crefname{table}{Table}{Tables}
\crefname{algorithm}{Algorithm}{Algorithms}
\crefname{equation}{Eq.}{Eqs.}

\begin{document}

\title{Co-PLNet: A Collaborative Point-Line Network for Prompt-Guided Wireframe Parsing}

\author{Chao Wang, Xuanying Li, Cheng Dai, Jinglei Feng, Yuxiang Luo, Hao Qin and Yuqi Ouyang
% --- 第一个脚注：投稿日期、基金支持、共同一作声明 ---
\thanks{Chao Wang and Xuanying Li contributed equally to this work. (Corresponding author: Yuqi Ouyang).}
% --- 第二个脚注：四川大学的作者（包括软件学院和计算机学院）---
\thanks{Chao Wang, Xuanying Li, and Cheng Dai are with the College of Software, Sichuan University, Chengdu 610065, China (email: \{2023141461095, 2023141461216\}@stu.scu.edu.cn; daicheng@scu.edu.cn).}
\thanks{Jinglei Feng is with the Department of Radiology, North Sichuan Medical College, Nanchong 637000, China (email: jingleifeng82@gmail.com).}
% --- 第三个脚注：其他单位的作者 ---
% \thanks{H. Huang is with the College of Electrical Engineering and Information, Northeast Agricultural University, Harbin 150030, China.}
% \thanks{Yuxiang Luo is with the Graduate School of Information, Production and Systems, Waseda University, Kitakyushu 808-0135, Japan (email: yuxiang.luo@ruri.waseda.jp).}
\thanks{Hao Qin is with the School of Electrical and Electronic Engineering, University College Dublin, Dublin, D04 V1W8 Ireland, and also with the School of Electronics and Information Engineering, Sichuan University, Chengdu, 610065, China (e-mail: hao.qin@scu.edu.cn).}
\thanks{Yuxiang Luo and Yuqi Ouyang are with the College of Computer Science, Sichuan University, Chengdu 610065, China (e-mail: yuxiang.luo@ruri.waseda.jp; yuqi.ouyang@scu.edu.cn).}}

\markboth{Journal of \LaTeX\ Class Files, Vol. 14, No. 8, August 2015}
{Shell \MakeLowercase{\textit{et al.}}: Bare Demo of IEEEtran.cls for IEEE Journals}
\maketitle

\begin{abstract}
Wireframe parsing aims to recover line segments and their junctions to form a structured geometric representation useful for downstream tasks such as Simultaneous Localization and Mapping (SLAM). Existing methods predict lines and junctions separately and reconcile them post-hoc, causing mismatches and reduced robustness. We present Co-PLNet, a point-line collaborative framework that exchanges spatial cues between the two tasks, where early detections are converted into spatial prompts via a Point-Line Prompt Encoder (PLP-Encoder), which encodes geometric attributes into compact and spatially aligned maps. A Cross-Guidance Line Decoder (CGL-Decoder) then refines predictions with sparse attention conditioned on complementary prompts, enforcing point-line consistency and efficiency. Experiments on Wireframe and YorkUrban show consistent improvements in accuracy and robustness, together with favorable real-time efficiency, demonstrating our effectiveness for structured geometry perception. 
Our code is available at \url{https://github.com/GalacticHogrider/Co-PLNet}.
\end{abstract}

\begin{IEEEkeywords}
Attention, Holistic Attraction Field Representation, Line Segment, Mutual Prompt, Wireframe Parsing
\end{IEEEkeywords}

\IEEEpeerreviewmaketitle

\section{Introduction}

\begin{figure}[!t]
    \centering
    \includegraphics[width=0.91\linewidth]{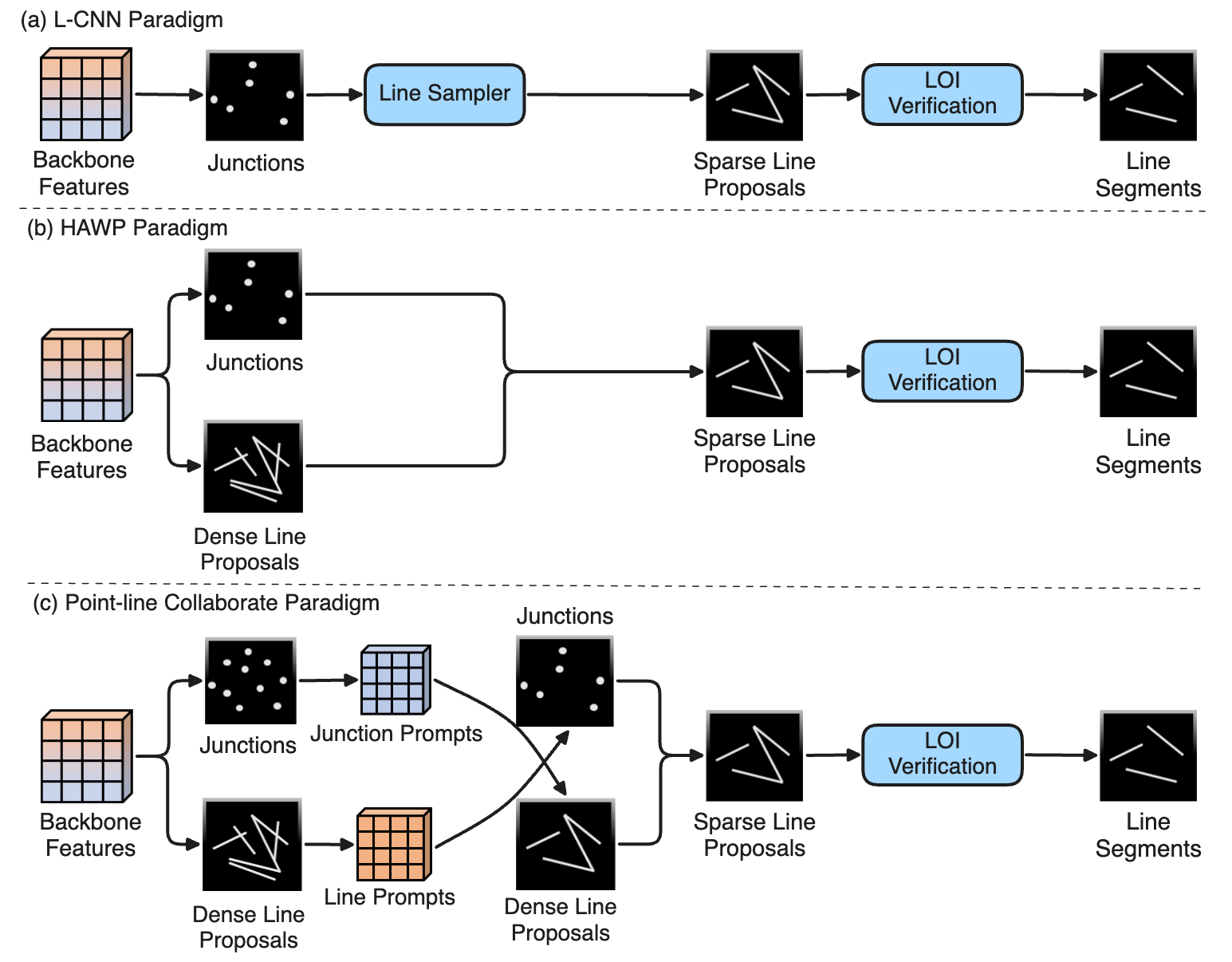}
    \caption{Conceptual comparison between existing wireframe parsing paradigms and our Co-PLNet. Note that this figure emphasizes architectural differences rather than visual output quality
    }
    \label{fig:paradigm}
\end{figure}

Wireframe parsing detects line segments and junctions to form structured geometric representations that support downstream SLAM \cite{airVo,xu2025airslam,liu2021pluckernet}, scene parsing \cite{wu2010learning,duan2015image}, and high-level visual recognition \cite{lazarow2022instance}, consistent with recent studies emphasizing fine-grained structural preservation in dense visual prediction \cite{li2026frequency,li2026position}.
 The interconnections between lines and junctions are important in wireframe parsing, while existing solutions generally utilize these interconnections in a limited way and are thus constrained in performance. As depicted in \cref{fig:paradigm}, traditional paradigms like L-CNN \cite{dai2022fully}\cite{xu2021line}\cite{zhou2019end} generate massive line proposals and filter them, which is computationally expensive. HAWP \cite{xue2020holistically}\cite{xue2023holistically} recasts sparse line proposals into dense predictions and aligns endpoints with junctions, improving efficiency and accuracy. PLNet \cite{xu2025airslam} builds on HAWP with cascaded U-Net \cite{ronneberger2015u} and deep supervision. However, since lines and junctions are predicted independently before post-hoc endpoint adjustments, this late alignment has limited ability to revise poorly generated candidates. This often leads to endpoint mismatches or incomplete structures in complex scenes, which adversely affects downstream SLAM tasks relying on their joint use.

\begin{figure*}[!t]
    \centering
    \includegraphics[width=0.90\linewidth]{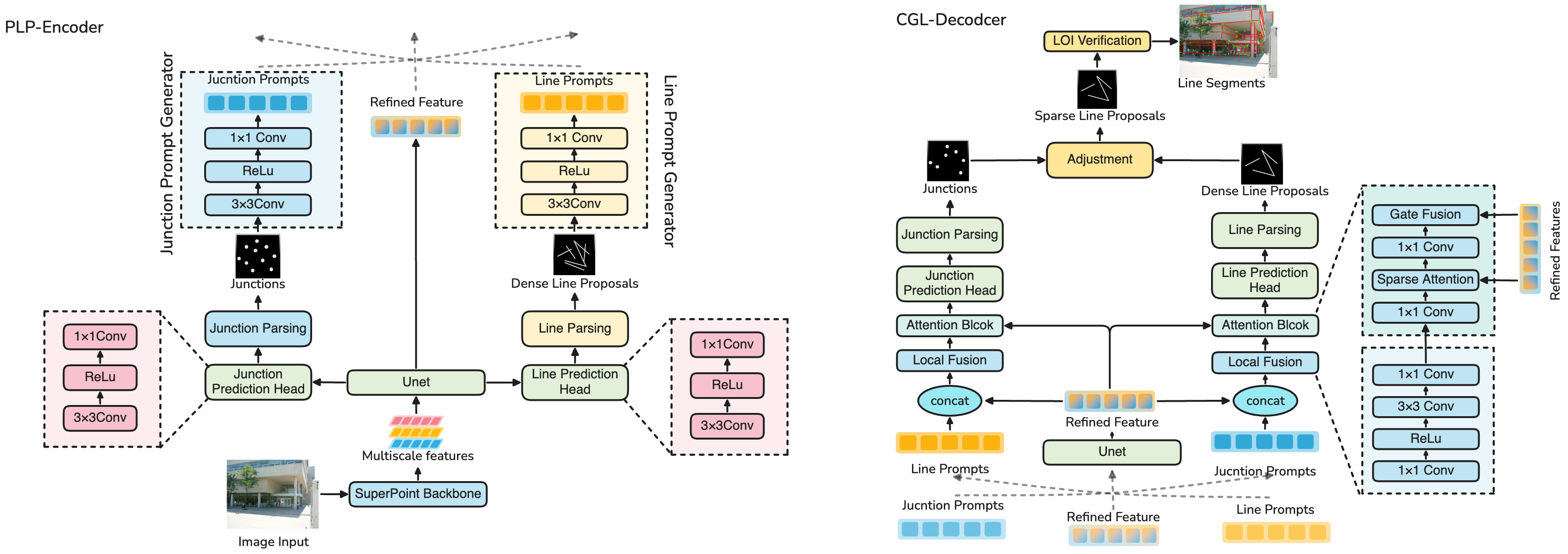}
    \caption{Overview of the Co-PLNet framework. The PLP-Encoder generates spatial prompts from junction and line predictions, which are refined by the CGL-Decoder to produce accurate line segment predictions.}
    \label{fig:framework}
\end{figure*}

Building upon the basic parsing components of HAWP and PLNet, we propose Co-PLNet, which introduces a novel intermediate point-line interaction stage for wireframe parsing. Depicted in \cref{fig:paradigm}, unlike traditional feature fusion or attention mechanisms lacking geometric supervision, our model performs early detection and directly converts explicit point-line predictions into spatial prompts exchanged across tasks, thereby ensuring point-line consistency. More specifically in \cref{fig:framework}, our Point-Line Prompt Encoder (PLP-Encoder) transforms geometric information into structured spatial prompts, supporting the prediction refinement in the Cross-Guidance Line Decoder (CGL-Decoder). Our main contributions are:

\begin{itemize}
    \item We propose a point-line collaborative paradigm that exchanges spatial prompts between junction and line detection, improving point-line consistency for wireframe parsing.
    \item We introduce a prompt-based encoding and decoding strategy that effectively fuses line and junction cues, enabling more robust and geometrically consistent predictions.
    \item Sparse attention is designed within cross-guided decoding to efficiently integrate complementary information, balancing accuracy and real-time performance.
    \item Comprehensive experiments on the Wireframe and YorkUrban datasets demonstrate our competitive performance.
\end{itemize}
% (i)We propose a point-line collaborative paradigm that exchanges spatial prompts between junction and line detection, improving point-line consistency for wireframe parsing.
% (ii)We design a Point-Line Prompt Encoder (PLP-Encoder) that converts geometric details of junctions and line segments into corresponding spatial prompts for downstream detection.
% (iii)We develop a Cross-Guidance Line Decoder (CGL-Decoder) that uses sparse attention to refine point and line predictions by integrating prompts from the complementary branch.
% (iv)Comprehensive experiments on the Wireframe and the YorkUrban datasets demonstrate the competitive performance of our model.

\section{Proposed Method}
\label{sec:format}
% We propose the Co-PLNet framework, which employs a novel point-line interactive wireframe parsing paradigm to achieve mutual optimization of line segment and junction detection. To this end, we design a PLP-encoder to encode the physical information of points and lines, and develop a CGL-decoder to fuse these prompts and generate predictions that provide more reliable and consistent point-line structures.

\subsection{Point-Line Collaborative Paradigm}
Existing parsers often use standard feature fusion or attention lacking direct supervision on junctions or endpoints, weakening geometric meanings. We propose a point-line collaborative paradigm that enables bidirectional interaction, yielding geometrically coherent predictions:

\begin{equation}
y = \text{LOI}(\hat{y}_J = f_J(Z, y_L^{(0)}), \hat{y}_L = f_L(Z, y_J^{(0)})),
\label{eq:Framework}
\end{equation}
where $Z$ denotes the refined feature map, $y_J^{(0)}$ and $y_L^{(0)}$ are the junction and line prompts generated by the PLP-Encoder, and $f_J$ and $f_L$ denote the junction and line prediction paths in the CGL-Decoder. In this formulation, line prompts are used to guide junction prediction, while junction prompts are used to guide line prediction, establishing the point-line interaction. 

\subsection{Point-Line Prompt Encoder}
Depicted in \cref{fig:framework}, the PLP-Encoder receives the input image and first applies SuperPoint~\cite{detone2018superpoint} to extract multi-scale features. These are aggregated by a U-Net adopted from PLNet~\cite{xu2025airslam} to obtain a refined feature representation. From this refined feature, two lightweight prediction heads together with parsing modules following HAWP~\cite{xue2020holistically} produce junction and dense line proposals, respectively. The goal of the PLP-Encoder is to transform these raw geometric predictions into spatial prompts that can guide the second-stage refinement in the CGL-Decoder.

Prior to junction parsing, we follow HAWP \cite{xue2020holistically} and use a point prediction head to predict a junction heatmap $H_{J}$ and pixel offsets $\Delta C_{J}$. Let $C$ denote all pixel coordinates in the image. The junction parsing starts by refining pixel coordinates with the offsets:

\begin{equation}
\label{eq:plp_junc1}
C_{J}\;=\; C + \Delta C_{J}.
\end{equation}

Then, each junction coordinate is assigned with a confidence value by normalizing the heatmap:
\begin{equation}
\label{eq:plp_junc2}
V_{J} \;=\; \mathrm{softmax}\!\big(H_{J}\big).
\end{equation}
    
After non-maximum suppression (NMS) on the heatmap $H_{J}$ and thresholding on confidence $V_{J}$, the junction parsing computes a sparse set of high-confidence junction coordinates $C_{I}\in\mathbb{R}^{H\times W\times2}$ as geometrically precise junctions, which are written into a zero-initialized map at their pixel locations to form dense maps for convolutional encoding.

For line parsing, we follow the HAFM \cite{xue2020holistically} and use the line prediction head to compute the per-pixel scalars $d,\theta,\theta_{1},\theta_{2}$ together with a small residual $r$ at every pixel location $c_{i}$. Here $d$ is the perpendicular distance from $c_{i}$ to the associated line segment; $\theta$ is the segment’s tangent orientation that rotates the image axes to the segment-aligned frame; $\theta_{1}$ and $\theta_{2}$ are endpoint angles measured in the segment-aligned frame between the perpendicular from $c_{i}$ to the segment and the rays pointing toward the two endpoints; $r$ refines $d$ for sub-pixel accuracy. Given these quantities, the two endpoints decoded at $c_{i}$ are
\begin{equation}
\label{eq:hafm_decode}
C_{ep}
=
(d+r)
\begin{pmatrix}
\cos\theta & -\sin\theta\\
\sin\theta & \cos\theta
\end{pmatrix}
\begin{pmatrix}
1 & 1\\
\tan\theta_{1} & \tan\theta_{2}
\end{pmatrix}
+\bigl[\,\mathbf c_{i},\,\mathbf c_{i}\,\bigr],
\end{equation}
where $C_{ep}\in\mathbb{R}^{2\times2}$ stores the two endpoint coordinates of the line segment decoded at $c_{i}$. Flattening and evaluating $C_{ep}$ over all pixel locations in the image produces $C_{L}\in\mathbb{R}^{H\times W\times 4}$, depicting all dense proposals of line segments.

As shown in \cref{fig:framework}, after the junction parsing and the line parsing, the junction prompt generator and the line prompt generator transform the junction and line proposals into junction prompts and line prompts, respectively, converting raw geometric predictions into spatial prompts, as follows:

\begin{equation}
\label{eq:plp_enc_j}
y_{J}^{(0)} \;=\; \mathrm{Conv}\!\Big(\mathrm{ReLU}\big(\mathrm{Conv}(C_{J}\big)\big)\Big),
\end{equation}
\begin{equation}
\label{eq:plp_enc_l}
y_{L}^{(0)} \;=\; \mathrm{Conv}\!\bigg(\mathrm{ReLU}\Big(\mathrm{Conv}(C_{L}\big)\Big)\bigg),
\end{equation}
where the kernel sizes of the two convolutions are depicted in \cref{fig:framework}. For all the channel settings, we refer to \cref{sec:ID}. The computations of $y_{J}^{(0)}$ and $y_{L}^{(0)}$ concludes the PLP-Encoder, resulting in spatially dense prompts that preserve junction density, local connectivity, and line orientation statistics, providing geometry-aware guidance for the refined predictions of line segments $y$ in the CGL-Decoder, as presented next.

\subsection{Cross-Guidance Line Decoder}
Depicted in \cref{fig:framework}, the CGL-Decoder first performs local feature fusion via channel-wise concatenation and convolutional operations of three spatially aligned inputs, they are: the refined feature map $Z$ produced by the U-Net adopted from PLNet \cite{xu2025airslam}, line prompts $y_L^{(0)}$, and the junction prompts $y_J^{(0)}$, resulting in locally fused features $\tilde{Z}_{L}$ and $\tilde{Z}_{J}$.

To capture non-local semantics efficiently, we apply sparse multi-head cross-attention over windowed features $\tilde{Z}_{L}$, $\tilde{Z}_{J}$, and $Z$. A $1{\times}1$ convolution $\psi(\cdot)$ first reduces channel dimensions, after which features are partitioned into spatial windows and projected into multiple attention heads. Cross-attention is then performed within corresponding windows across the three feature sources:

\begin{equation}
    \bar{Z}_{L} = \text{MHA}\big(\psi(\tilde{Z}_{L}),\,\psi(Z),\,\psi(Z)\big),
\end{equation}

\begin{equation}
    \bar{Z}_{J} = \text{MHA}\big(\psi(\tilde{Z}_{J}),\,\psi(Z),\,\psi(Z)\big),
\end{equation}
where attended features $\bar{Z}_{L}$ and $\bar{Z}_{J}$ are produced, which are further injected into the refined representation via gated residual fusion to suppress noise:

% \begin{equation}
%     Z_{L}' = Z + G_{L} \odot \bar{Z}_{L},
% \end{equation}

% \begin{equation}
%     Z_{J}' = Z + G_{J} \odot \bar{Z}_{J},
% \end{equation}
\begin{equation}
    Z_{L}' = Z + G_{L} \odot \bar{Z}_{L},\quad
    Z_{J}' = Z + G_{J} \odot \bar{Z}_{J},
\end{equation}
where $\odot$ denotes element-wise multiplication, $G_{L}$ and $G_{J}$ are learnable gating masks. The refined features $Z_L'$ and $Z_J'$ are used for line and junction prediction, respectively. 
Prediction heads and parsing modules similar to those in the PLP-Encoder are applied to obtain dense line proposals $y_L$ and junctions $y_J$. 
Following HAWP \cite{xu2025airslam}, line endpoints are associated with nearby junctions within a certain threshold and duplicates are removed, yielding sparse line proposals $y$. The mutual guidance between line and junction prompts is maintained throughout prediction, enforcing point-line consistency as formalized in \cref{eq:Framework}. Finally, Line-of-Interest (LOI) verification is applied to $y$: features are sampled along each line and scored by an MLP, while the top-$k$ proposals are retained as final predictions. Except the Superpoint module, all components in our solution are trained end-to-end by the following loss:
 
\begin{equation}
    \mathcal{L}
=\sum_{m\in\{\text{PLP},\text{CGL}\}}\!\Big(
\mathcal{L}_{\mathrm{line}}^{(m)}
+\mathcal{L}_{\mathrm{junc}}^{(m)}
+\mathcal{L}_{\mathrm{aux}}^{(m)}
\Big)
+\mathcal{L}_{\mathrm{LOI}},
\end{equation}
where $\mathcal{L}_{\mathrm{line}}$ supervises the line parameter maps $(d,\theta,\theta_{1},\theta_{2},r)$;
$\mathcal{L}_{\mathrm{junc}}$ supervises the junction heatmap and offsets $(H_J,\Delta C_J)$; and $\mathcal{L}_{\mathrm{aux}}$ enforces geometric consistency between dense line proposals $C_L$ and the target geometry.
The index $m$ denotes the PLP-Encoder or the CGL-Decoder.
$\mathcal{L}_{\mathrm{LOI}}$ supervises line confidence prediction by aggregating features along each proposal $y$. Due to space limitations, the definitions of $\mathcal{L}_{\mathrm{line}}$, $\mathcal{L}_{\mathrm{junc}}$, and $\mathcal{L}_{\mathrm{aux}}$ are not presented here, detailed formulations can be found in \cite{xu2025airslam}.

\section{Experiments}
\begin{figure*}[!t]
    \centering
    \includegraphics[width=0.70\linewidth]{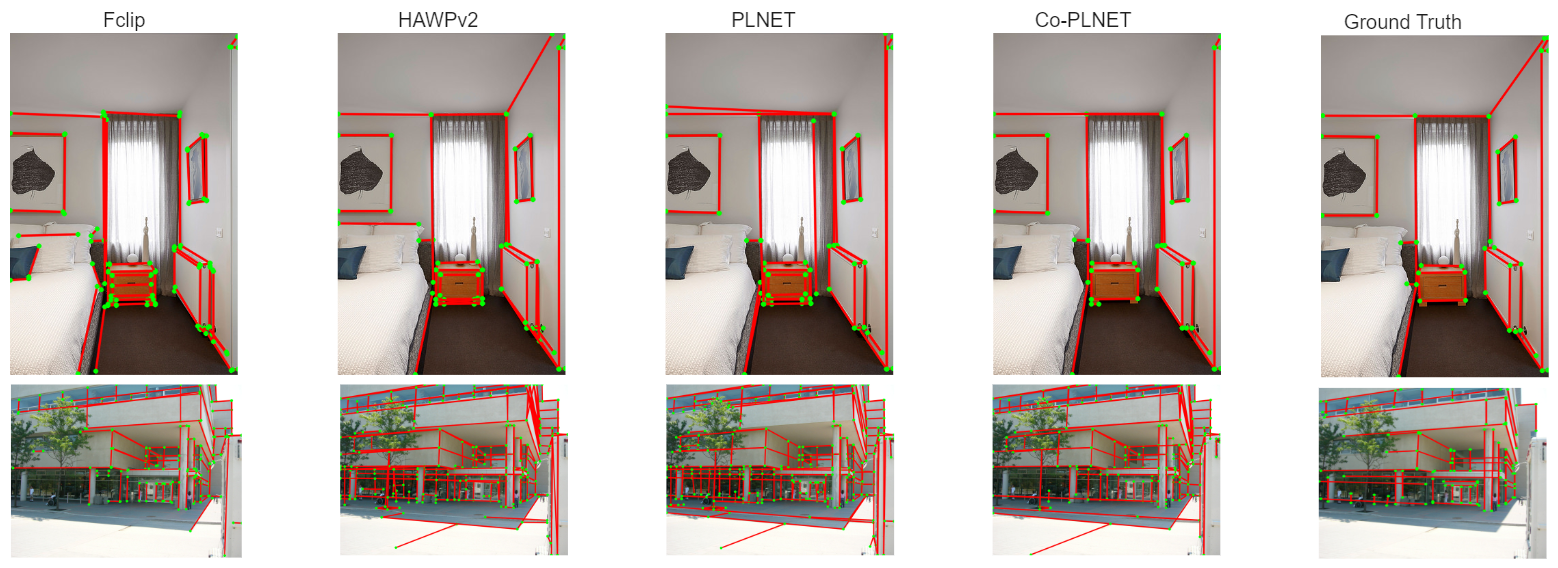}
    \caption{Visualization of parsing results. The first and second rows show images from the Wireframe and YorkUrban datasets, respectively, with predicted wireframes aligned to ground-truth annotations.}
    \label{fig:placeholder}
\end{figure*}

\begin{table}[!t]
\centering
\caption{Performance comparison on the Wireframe and YorkUrban datasets. Results are reported for color inputs, while methods marked with \textsuperscript{*} correspond to grayscale inputs. The highest result is in \textbf{bold}, and the second highest is \underline{underlined}}
%\vspace{0.2cm}
\label{tab:performance_comparison_small}
\resizebox{0.43\textwidth}{!}{%
\begin{tabular}{l|ccc|ccc|l}
\hline
\multicolumn{1}{l|}{\multirow{2}{*}[-0.6ex]{Methods}} &
\multicolumn{3}{c|}{Wireframe} &
\multicolumn{3}{c|}{YorkUrban} &
\multicolumn{1}{l}{\multirow{2}{*}[-0.6ex]{FPS}} \\
\hhline{~|---|---|~} % ← 用 hhline 让分割线与竖线“拉通”
& \rule{0pt}{2.2ex} sAP$^5$ & sAP$^{10}$ & sAP$^{15}$ &
  sAP$^5$ & sAP$^{10}$ & sAP$^{15}$ & \\
\hline
% =========== Color Input Group ===========
\textmd{AFM\cite{xue2019learning}}   & 18.5 & 24.4 & 27.5 & 7.3 & 9.4 & 11.1 & 10.4 \\
\textmd{AFM++\cite{xue2019learning}} & 27.7 & 32.4 & 34.8 & 9.5 & 11.6 & 13.2 & 8.0 \\
\textmd{L-CNN\cite{zhou2019end}} & 59.7 & 63.6 & 65.3 & 25.0 & 27.1 & 28.3 & 29.6 \\
\textmd{LETR\cite{xu2021line}}  & 59.2 & 65.2 & 67.7 & 23.9 & 27.6 & 29.7 & 2.0 \\
\textmd{F-Clip\cite{dai2022fully}} & 64.3 & 68.3 & 69.1 & 28.6 & 31.0 & 32.4 & \underline{82.3} \\
\textmd{ELSD\cite{zhang2021elsd}}  & 64.3 & 68.9 & 70.9 & 27.6 & 30.2 & 31.8 & 42.6 \\
\textmd{HAWPv2\cite{xue2023holistically}} & \underline{65.7} & \underline{69.7} & \underline{71.3} & \underline{28.9} & \underline{31.2} & \underline{32.6} & \textbf{85.2} \\
\textmd{Co-PLNet} & \textbf{68.4} & \textbf{72.3} & \textbf{73.8} & \textbf{32.7} & \textbf{35.6} & \textbf{36.6} & 76.8 \\
\hline
% =========== Grayscale Input Group ===========
\textsuperscript{*}HAWPv2\cite{xue2023holistically} & 63.6 & 67.7 & 69.5 & 26.6 & 29.0 & 30.3 & \textbf{85.2} \\
\textsuperscript{*}PLNet\cite{xu2025airslam}  & \underline{65.2} & \underline{69.2} & \underline{70.9} & \underline{29.3} & \underline{32.0} & \underline{33.5} & \underline{79.4} \\
\textsuperscript{*}Co-PLNet & \textbf{67.9} & \textbf{71.7} & \textbf{73.3} & \textbf{32.3} & \textbf{35.3} & \textbf{36.4} & 76.8 \\
\hline 
\end{tabular}%
}
%\vspace{2pt}
\end{table}
% \label{sec:pagestyle}
\subsection{Datasets and Evaluation Metrics}
Our model is evaluated on the Wireframe \cite{huang2018learning} and YorkUrban \cite{denis2008efficient} datasets. The Wireframe dataset contains 5,000 training images and 462 test images with richly annotated line segments and junctions from diverse built environments. 
The YorkUrban dataset includes 102 urban-scene test images and is commonly used to evaluate cross-domain generalization.

We evaluate detection accuracy by structural average precision (sAP), and runtime efficiency using frames per second (FPS). A predicted segment with endpoints $(\hat{c}_{1}, \hat{c}_{2})$ is considered a true positive if it matches a ground-truth segment with fixed endpoints $(c_{1}, c_{2})$ satisfying:

\begin{equation}
\bigl\|\hat{c}_{1}-c_{1}\bigr\|_2^{2}
+\bigl\|\hat{c}_{2}-c_{2}\bigr\|_2^{2}
\le l.
\end{equation}
Following \cite{zhou2019end}, we evaluate at thresholds $l\in\{5,10,15\}$ and report $\mathrm{sAP}^{5}$, $\mathrm{sAP}^{10}$, and $\mathrm{sAP}^{15}$. To quantify point–line consistency, we report the endpoint mismatch rate, defined as the percentage of predicted endpoints without a detected junction within 15 pixels.

\subsection{Implementation Details} \label{sec:ID}
All images are resized to $512\times512$. Experiments are conducted on an NVIDIA RTX~4080 GPU using PyTorch. The Superpoint module is fixed with the pre-trained version in \cite{detone2018superpoint}. Following PLNet~\cite{xu2025airslam}, we adopt consistent configurations for the backbone, proposal generation, and LOI verification, including 256-channel U-Net features, a 0.008 junction threshold, a 10-pixel proposal range, and the retention of the top-$1000$ proposals. To balance structural accuracy with real-time efficiency, we employ lightweight configurations for the collaborative modules: both junction and line prompts are encoded with 16 channels, and the sparse multi-head cross-attention projects features to 32 channels using 4 heads with a window size of 8.
For training, Adam \cite{adam2014method} is used with a learning rate of $4\times10^{-4}$ for 35 epochs, reduced to $4\times10^{-5}$ for the final 5 epochs. The batch size is set to 6.

%\subsection{Evaluation Metrics}
\subsection{Comparison with SOTA Methods}
To comprehensively evaluate our performance, we conducted extensive comparisons with state-of-the-art (SOTA) approaches on the two datasets. Tabulated in \cref{tab:performance_comparison_small}, for both color and greyscale inputs, our method consistently outperforms prior work across all sAP thresholds on both datasets, while maintaining real-time efficiency at 76.8 FPS with only a marginal parameter overhead compared with PLNet (8.89M vs. 8.78M). Visual comparisons in \cref{fig:placeholder} highlight the improvements of our framework. Compared with F-Clip, HAWPv2, and PLNet, Co-PLNet produces more complete and geometrically consistent line–junction structures, closely matching the ground truth annotations. These findings establish Co-PLNet as a competitive approach for accurate and efficient wireframe parsing.

\subsection{Ablation Study}
We evaluate different model components on Wireframe and YorkUrban with greyscale inputs. As summarized in \cref{tab:ablation}, the baseline achieves sAP scores of 65.2, 69.2, 70.9 on Wireframe and 29.3, 32.0, 33.5 on YorkUrban. Introducing the point-to-line (PL) prompt improves line-proposal quality and point–line connectivity, enhancing accuracy especially at sAP$^{15}$, while the line-to-point (LP) prompt focuses on refining junction localization, therefore limited at boosting sAP$^{5}$. Enabling both prompts jointly increases precision further and reduces endpoint mismatch (Mis.) from 12.4 to 9.6 on Wireframe, demonstrating the complementary benefits of mutual point–line prompting. Furthermore, replacing SuperPoint with a standard CNN increases endpoint mismatch from 7.8 to 10.2 on Wireframe and from 8.5 to 11.3 on YorkUrban, while adding SA improves sAP$^{15}$ from 72.6 to 73.3 and from 35.1 to 36.4, respectively; dense attention further reaches 73.6 and 36.7 but drops FPS from 76.8 to 42.1, justifying the sparse design. Overall, these results highlight that each component contributes to more accurate junctions and better line connectivity while keeping the system real-time.
\begin{table}[t]
\centering
\scriptsize
\setlength\tabcolsep{2pt}
\renewcommand{\arraystretch}{0.95}
\caption{Ablated performance under different system configurations. SP denotes the fixed SuperPoint; omitting it implies using a standard CNN. PL and LP denote point-to-line and line-to-point prompts; LF is local feature fusion; SA is sparse attention; DA is dense attention. Mis. indicates the endpoint mismatch rate (\%).}
\label{tab:ablation}
\resizebox{0.43\textwidth}{!}{%
\begin{tabular}{cccccc cccc cccc c} % 总共 15 列
\toprule
\multirow{2}{*}[-0.8ex]{SP} &
\multirow{2}{*}[-0.8ex]{PL} &
\multirow{2}{*}[-0.8ex]{LP} &
\multirow{2}{*}[-0.8ex]{LF} &
\multirow{2}{*}[-0.8ex]{SA} &
\multirow{2}{*}[-0.8ex]{DA} & 
\multicolumn{4}{c}{Wireframe} &
\multicolumn{4}{c}{YorkUrban} &
\multirow{2}{*}[-0.6ex]{FPS} \\
\cmidrule(lr){7-10}\cmidrule(lr){11-14} % 跨列参数已自适应修改
& & & & & &
\rule{0pt}{2.2ex} sAP$^{5}$ & sAP$^{10}$ & sAP$^{15}$ & Mis.$\downarrow$ &
sAP$^{5}$ & sAP$^{10}$ & sAP$^{15}$ & Mis.$\downarrow$ & \\
\midrule
\ding{51} & & & & & & 65.2 & 69.2 & 70.9 & 12.4 & 29.3 & 32.0 & 33.5 & 13.1 & 79.4 \\
\ding{51} & \ding{51} & & \ding{51} & & & 66.1 & 70.3 & 72.3 & 11.2 & 30.1 & 32.7 & 34.8 & 12.0 & 78.5 \\
\ding{51} & & \ding{51} & \ding{51} & & & 66.8 & 70.4 & 71.5 & 10.8 & 30.8 & 32.5 & 33.9 & 11.5 & 78.3 \\
\ding{51} & \ding{51} & \ding{51} & \ding{51} & & & 67.1 & 71.2 & 72.6 & 9.6 & 31.4 & 33.9 & 35.1 & 10.2 & 77.6 \\
 & \ding{51} & \ding{51} & \ding{51} & \ding{51} &  & 66.7 & 70.9 & 72.1 & 10.2 & 31.1 & 34.2 & 35.0 & 11.3 & 77.9 \\
\ding{51} & \ding{51} & \ding{51} & \ding{51} & \ding{51} & & 67.9 & 71.7 & 73.3 & 7.8 & 32.3 & 35.3 & 36.4 & 8.5 & 76.8 \\
\ding{51} & \ding{51} & \ding{51} & \ding{51} & & \ding{51} & 68.1 & 71.9 & 73.6 & 7.5 & 32.5 & 35.5 & 36.7 & 8.2 & 42.1 \\
\bottomrule
\end{tabular}%
}
\end{table}
\subsection{Parameter Study}

\cref{tab:window_sweep} investigates the impact of window size for sparse attention, using Wireframe and YorkUrban images converted to greyscale. All scoring metrics follow a consistent trend, peaking at a window size of 8 across both datasets. A smaller window size of 4 increases inference speed but reduces precision, while increasing the window size to 16 does not improve accuracy and adds computational overhead. These observations justify 8 as the optimal window size in sparse attention for balancing accuracy and efficiency.

\begin{table}[!t]
\centering
\small
\setlength\tabcolsep{4pt} % 稍微缩小列间距以适应更多列
\caption{Performance impact of window size $w$ for sparse attention.}
\label{tab:window_sweep}
\resizebox{0.38\textwidth}{!}{%
\begin{tabular}{c ccc ccc c}
\toprule
\multirow{2}{*}{\makecell{Window \\ Size}} & \multicolumn{3}{c}{Wireframe} & \multicolumn{3}{c}{YorkUrban} & \multirow{2}{*}{FPS} \\
\cmidrule(lr){2-4} \cmidrule(lr){5-7}
& sAP$^{5}$ & sAP$^{10}$ & sAP$^{15}$ & sAP$^{5}$ & sAP$^{10}$ & sAP$^{15}$ & \\
\midrule
4  & 67.2 & 71.3 & 72.8 & 31.1 & 34.7 & 35.8 & \textbf{78.4} \\
8  & \textbf{67.9} & \textbf{71.7} & \textbf{73.3} & \textbf{32.3} & \textbf{35.3} & \textbf{36.4} & 76.8 \\
16 & 67.6 & 71.5 & 73.1 & 31.9 & 35.1 & 36.1 & 74.2 \\
\bottomrule
\end{tabular}
}
\end{table}
\section{Conclusion}
We present Co-PLNet, a prompt-based collaborative framework for wireframe parsing that tightly integrates line and junction prediction. Preliminary junction and line predictions are converted into spatial prompts by the PLP-Encoder and exchanged in the CGL-Decoder via sparse attention, enhancing point–line consistency and achieving competitive structural average precision on the Wireframe and YorkUrban datasets, while operating at 76.8 FPS. Ablation studies further confirm the effectiveness of each model component, particularly the spatial prompts that provide mutual guidance.

\vfill\pagebreak

% \section{REFERENCES}
% \label{sec:refs}

% List and number all bibliographical references at the end of the
% paper. The references can be numbered in alphabetic order or in
% order of appearance in the document. When referring to them in
% the text, type the corresponding reference number in square
% brackets as shown at the end of this sentence \cite{C2}. An
% additional final page (the fifth page, in most cases) is
% allowed, but must contain only references to the prior
% literature.

% Please follow the IEEE Citation Guidelines, \url{https://ieee-dataport.org/sites/default/files/analysis/27/IEEE\%20Citation\%20Guidelines.pdf} for formatting of references.

\bibliographystyle{IEEEtran}
\bibliography{refs}

\end{document}